\documentclass[10pt,twocolumn]{article}

\usepackage{amsmath}
\usepackage[a4paper,
            top=0.70in,
            bottom=0.75in,
            left=0.70in,
            right=0.70in,
            columnsep=0.25in]{geometry}
\usepackage{times}
\usepackage{microtype}
\usepackage{amsmath,amssymb,amsfonts}
\usepackage{booktabs}
\usepackage{graphicx}
\usepackage{adjustbox}
\usepackage{xcolor}
\usepackage[numbers,sort&compress]{natbib}
\usepackage[colorlinks=true,
            citecolor=blue,
            linkcolor=blue,
            urlcolor=blue]{hyperref}
\usepackage{enumitem}
\setlist{nosep}
\usepackage[utf8]{inputenc} 
\usepackage[T1]{fontenc}    
\usepackage{hyperref}       
\usepackage{url}            
\usepackage{booktabs}       
\usepackage{amsfonts}       
\usepackage{nicefrac}       
\usepackage{microtype}      
\usepackage{xcolor}         
\usepackage{booktabs}
\usepackage{adjustbox}
\usepackage{graphicx}
\usepackage{natbib}

\newcommand{\clocal}{c_{\mathrm{local}}}
\newcommand{\cglobal}{c_{\mathrm{global}}}

\title{When Confidence Signals Disagree: Local and Global Confidence in Autoregressive Language Model}

\author{%
  Julio C. Amador D\'iaz L\'opez\thanks{
  SiftyML, LTD\\
  London, United Kingdom \\
  \texttt{julio@siftyml.com} \\
  Code available at \texttt{https://github.com/julioadl/stochastic-parrots}
}}

\begin{document}

\maketitle

\begin{abstract}

Modern predictive systems expose multiple quantities that are commonly
interpreted as measures of confidence. However, these quantities can summarize
different aspects of the predictive process. This distinction matters when
confidence is used to evaluate reliability or inform downstream oversight and
control. We investigate whether different confidence readouts are empirically
interchangeable in an autoregressive language model by comparing local
confidence, defined from the probability of the greedy-selected answer token,
with global confidence, defined from modal-answer frequency under repeated
sampling. Across MMLU and ARC Challenge, the two signals are weakly correlated
($r=0.086$ and $r=0.175$, respectively) and differ substantially in their
association with correctness: global confidence is moderately associated with
correctness ($r=0.494$ and $r=0.374$), whereas local confidence shows little
association ($r=0.102$ and $r=0.072$). We further test whether question-level
disagreement between the signals is associated with sampling instability. On
ARC, larger local--global confidence gaps are associated with higher answer
entropy (Spearman $\rho=0.42$), more distinct sampled answers ($\rho=0.41$),
and lower modal-answer concentration ($\rho=-0.42$). The gap--entropy
association persists when disagreement and instability are estimated from
disjoint stochastic samples, indicating that it is not explained by shared
finite-sample variation. The corresponding relationship is substantially
weaker on MMLU, where only 4\% of questions exhibit sampling instability.
These results show that confidence readouts derived from the same predictive
system are not empirically interchangeable and that their disagreement can
provide a diagnostic of unstable sampling behavior. Confidence should
therefore be treated as an explicitly defined measurement rather than as a
single intrinsic scalar property of a model, particularly when it is used to
inform downstream evaluation, oversight, or control.

\end{abstract}
\section{Introduction}
\label{sec:introduction}

Large language models increasingly participate in systems that must decide not
only \emph{what} to predict, but also \emph{how much confidence} to place in
that prediction. Confidence estimates are used to evaluate calibration,
trigger abstention or human review, weight feedback, support selective
prediction, and characterize the reliability of aligned models
\citep{tian2023calibration}.
In these settings, confidence functions as an interface between the model's
predictive behavior and downstream mechanisms for oversight and control.

Yet ``model confidence'' is not a uniquely defined quantity. Modern language
models expose several signals that are routinely interpreted as confidence,
including token probabilities, sampled-answer frequencies, verbalized
confidence, and other derived uncertainty estimates. Prior work has shown that
such signals can differ substantially in calibration and reliability
\citep{tian2023calibration,kumar2024confidence}. This raises a more basic
measurement question: when different readouts of the same predictive system
are called confidence, to what extent do they actually describe the same
behavioral state?

We study this question in a deliberately simple setting. In an autoregressive
language model, a \emph{local} confidence signal can be obtained from the
conditional probability assigned to a selected answer token, whereas a
\emph{global} confidence signal can be estimated from the concentration of
answers produced under repeated sampling. These quantities are statistically
distinct by construction: one summarizes a particular conditional decision,
while the other summarizes the induced distribution over complete sampled
answers. There is therefore no theoretical requirement that they coincide.
The empirical questions are instead how large that mismatch is in practice, whether the two
signals carry equivalent information about correctness, and whether their
disagreement is systematically related to observable instability in model
behavior.

These questions matter beyond the choice of uncertainty metric. Alignment and
reliability procedures increasingly rely on confidence as a signal for
deciding when a model should be trusted, corrected, deferred, or assigned
greater or lesser weight. If confidence depends materially on how the
predictive distribution is interrogated, then downstream procedures may be
conditioning on different properties of the same model while treating them as
a common quantity. Before confidence can be aligned with correctness,
preferences, human judgments, or external feedback, its operational meaning
must therefore be made explicit.

We investigate this issue using a locally served Llama~3.3 70B model. The primary local--global
comparison uses MMLU and ARC, whose constrained answer spaces allow
token-level probabilities and sampled-answer frequencies to be compared at a
common answer level. Local confidence is defined as the probability assigned
to the greedy-selected answer token; global confidence is the empirical
frequency of the modal answer under repeated sampling; and their disagreement
is measured at the question level by the absolute difference between these
signals.

The experiments reveal three main patterns. First, local and global confidence
show weak empirical agreement on both MMLU and ARC, with non-trivial
confidence gaps occurring for a substantial fraction of questions. Second,
the signals differ markedly in their association with correctness: sampled
answer concentration is moderately associated with correctness, whereas local
token confidence shows little association. Third, on ARC, larger local--global
confidence gaps are systematically associated with greater sampling
instability, including higher answer entropy, more distinct sampled answers,
and lower modal-answer concentration. This relationship persists when
confidence disagreement and sampling instability are estimated from disjoint
stochastic samples, indicating that it is not explained by shared
finite-sample variation. The corresponding relationship is substantially
weaker on MMLU, where sampling instability is rare.

These findings shift the focus from whether different confidence signals
\emph{can} disagree, which follows from their definitions, to whether that
disagreement is empirically consequential. We make three contributions:
\begin{enumerate}
    \item We provide a controlled empirical characterization of local token
    confidence and global sampled-answer concentration as distinct confidence
    measurements within the same autoregressive model.
    \item We show that these signals exhibit weak agreement and substantially
    different associations with answer correctness.
    \item We directly link question-level confidence disagreement to sampling
    instability, finding a consistent association on ARC that persists under
    disjoint-sample estimation, alongside a substantially weaker relationship
    in the low-instability MMLU regime.
\end{enumerate}

More broadly, the results suggest that confidence should be treated as an
explicitly defined measurement of model behavior rather than as a single
intrinsic scalar property. This distinction becomes especially important when
confidence is used as an input to evaluation, oversight, or alignment
procedures: disagreement between confidence readouts implies that the choice of
measurement can itself determine what downstream systems infer about the
model's uncertainty.

\section{Related Work}
\label{sec:related}

\paragraph{Calibration and confidence in neural networks.}
A large literature studies whether predictive confidence corresponds to
empirical correctness. \citet{niculescu2005predicting} examine methods for
obtaining reliable probability estimates from supervised classifiers, while
\citet{guo2017calibration} show that modern neural networks can be poorly
calibrated and demonstrate that post-hoc temperature scaling can substantially
improve calibration. These approaches generally begin with a specified scalar
confidence score and ask whether that score is calibrated. Our work addresses
a complementary question: when multiple quantities derived from the same
predictive process can reasonably be interpreted as confidence, do they provide
empirically interchangeable measurements? In language models,
\citet{jiang2021know} show that probability-based confidence can vary in
reliability across question types and model scales, while
\citet{zhao2023slic} study the relationship between sequence likelihood and
generation quality. We focus specifically on the relationship between a local
token-probability readout and the concentration of the answer distribution
induced by repeated sampling.

\paragraph{Uncertainty and confidence in language models.}
Language models admit several operationalizations of confidence beyond token
probabilities. \citet{kadavath2022language} study model self-evaluation,
\citet{lin2022teaching} train models to express uncertainty in words, and
\citet{xiong2024llms} examine different methods for eliciting confidence from
LLMs. \citet{tian2023calibration} further show that confidence elicitation and
calibration depend on how confidence is obtained from models fine-tuned with
human feedback. More directly related to the present measurement question,
\citet{kumar2024confidence} compare verbalized confidence with token-level
probabilities and find that the two can be weakly aligned. Together, these
results demonstrate that different confidence readouts need not behave
equivalently. We extend this comparison to two quantities derived directly
from the predictive process: the probability of the greedy-selected answer
token and the empirical concentration of answers under repeated sampling.

\paragraph{Sampling-based uncertainty and self-consistency.}
Repeated generation provides an alternative view of model uncertainty by
characterizing the distribution of outputs produced for the same input.
\citet{malinin2021uncertainty} develop uncertainty measures for autoregressive
structured prediction based on sequence-level predictive distributions, while
\citet{kuhn2023semantic} address an important complication in open-ended
language generation by grouping semantically equivalent responses before
computing entropy. Our use of constrained multiple-choice answer spaces avoids
this additional semantic-equivalence problem and permits sampled answer
concentration to be estimated directly.

Sampling distributions are also used for inference. In particular,
\citet{wang2023selfconsistency} show that aggregating multiple sampled
reasoning paths through majority voting can improve reasoning accuracy, and
\citet{chen2023universal} extend self-consistency methods to more general
generation settings. These studies demonstrate that repeated samples contain
behaviorally useful information beyond a single generated output. Our focus
is different: we use the sampled answer distribution as a confidence readout
and ask how it relates to token-level confidence and question-level sampling
instability.

\paragraph{Confidence, elicitation, and downstream decision making.}
The distinction between confidence readouts becomes consequential when
confidence is used by downstream systems. Prompt formulation and elicitation
can materially affect model probabilities and outputs
\citep{zhao2021calibrate,sclar2024quantifying}, while internal representations
can contain information that is not straightforwardly reflected in surface
responses \citep{burns2023discovering,marks2023geometry}. These findings
motivate caution in treating any single observable confidence signal as a
complete description of the model's predictive state.

This issue also arises in deployed decision pipelines. For example,
\citet{ren2023robots} use repeated LLM outputs with conformal prediction for
robotic task planning rather than relying directly on token-level
probabilities. More generally, when confidence is used for abstention,
deferral, feedback, or control, the operational definition of confidence
determines what information is supplied to the downstream mechanism. Our
study isolates this measurement problem by directly comparing local token
confidence with sampling-derived answer concentration and testing whether
their disagreement is associated with observable sampling instability.

\section{Local and Global Confidence}
\label{sec:localglobal}

Let $x \in \mathcal{X}$ denote an input and $y^* \in \mathcal{Y}$ the corresponding ground-truth output. An autoregressive model induces a distribution over output sequences,
\begin{equation}
\hat{P}(y \mid x)
=
\prod_{t=1}^{T}
\hat{P}(y_t \mid y_{<t},x).
\end{equation}

This predictive distribution can support different scalar quantities that may be interpreted as confidence. We focus on two such quantities: a \emph{local} confidence signal derived from a conditional prediction at a particular generation step, and a \emph{global} confidence signal derived from the distribution over complete outputs.

\paragraph{Local confidence.}
Local confidence is derived from a conditional probability
\begin{equation}
c_{local}=\hat{P}(y_t \mid y_{<t},x),
\end{equation}
and therefore characterizes the concentration of probability at an individual autoregressive decision point. In the experiments below, we operationalize this quantity using the probability assigned to the greedy-selected answer token.

\paragraph{Global confidence.}
Global confidence characterizes concentration in the induced distribution over complete outputs. For an output-level answer $a$, this distribution can be approximated through repeated sampling from the model. We operationalize global confidence as the empirical probability of the modal answer,
\begin{equation}
c_{\mathrm{global}}(x)
=
\max_a \hat{P}(a \mid x),
\end{equation}
where $\hat{P}(a \mid x)$ denotes the empirical answer distribution obtained from repeated samples.

\paragraph{Local--global disagreement.}
Local and global confidence summarize different aspects of the same predictive process. The former measures probability concentration at a particular conditional decision, whereas the latter measures concentration after complete generated outputs are mapped to answers. There is therefore no requirement that the two quantities coincide.

For each input, we quantify their disagreement by the absolute confidence gap
\begin{equation}
D(x)
=
\left|
c_{\mathrm{local}}(x)
-
c_{\mathrm{global}}(x)
\right|.
\end{equation}

The existence of such disagreement is not itself surprising: local conditional probabilities and output-level concentration are distinct statistical quantities. The empirical questions are instead how strongly these confidence signals agree in practice, whether they exhibit similar relationships with correctness, and whether the magnitude of their disagreement is associated with variability in the model's sampled outputs.

For the latter comparison, we define correctness as
\begin{equation}
z(x)
=
\mathbb{I}\!\left(\hat{y}(x)=y^*\right).
\end{equation}
We treat the relationship between each confidence signal and $z(x)$ as a confidence--correctness association rather than as a sufficient characterization of probabilistic calibration.

\section{Sampling Instability}

The preceding section defines local and global confidence as distinct summaries of an autoregressive predictive process. We now consider whether disagreement between these confidence signals is associated with variability in the model's sampled outputs.

\subsection{Repeated Sampling and Output Variability}

For a fixed input $x$ and sampling configuration, repeated generation induces an empirical distribution over answers,
\begin{equation}
\hat{P}(a \mid x)
=
\frac{1}{N}
\sum_{i=1}^{N}
\mathbb{I}(a_i=a),
\end{equation}
where $a_i$ denotes the answer obtained from the $i$th sampled output.

The concentration of this distribution provides a direct measure of sampling stability. If repeated samples consistently produce the same answer, the empirical distribution is concentrated and the model is stable for that input under the specified sampling configuration. If probability mass is distributed across multiple answers, the input exhibits greater sampling instability.

\subsection{Measures of Sampling Instability}

We characterize sampling instability using complementary properties of the empirical answer distribution. These include its Shannon entropy,
\begin{equation}
H(x)
=
-\sum_a
\hat{P}(a \mid x)
\log \hat{P}(a \mid x),
\end{equation}
the number of distinct answers observed across repeated samples, and the empirical probability of the modal answer. We additionally define an input as unstable when repeated sampling produces more than one distinct answer.

These measures capture different aspects of the same phenomenon: entropy measures dispersion across answers, distinct-answer count measures support size, and modal-answer probability measures concentration on the most frequently generated answer.

\subsection{Confidence Disagreement and Sampling Instability}

The local--global confidence gap
\begin{equation}
D(x)
=
\left|
c_{\mathrm{local}}(x)
-
c_{\mathrm{global}}(x)
\right|
\end{equation}
provides an input-level measure of disagreement between the two confidence signals. Sampling instability provides a separate characterization of variability in the model's generated answers.

We test whether these quantities are systematically associated at the question level. Specifically, we examine whether larger local--global confidence gaps are associated with greater answer entropy, a larger number of distinct answers, lower modal-answer probability, and a greater likelihood of producing more than one answer under repeated sampling.

This analysis is associative rather than causal. In particular, $c_{\mathrm{global}}(x)$ and the sampling-instability measures are estimated from repeated outputs of the same predictive process and are therefore statistically related by construction. The analysis asks whether local--global confidence disagreement tracks observable sampling instability; it does not establish whether disagreement produces instability, instability produces disagreement, or both arise from properties of the underlying predictive distribution.

\section{Experimental Setup}
\label{sec:setup}

We investigate three empirical questions: (1) how strongly local and global
confidence agree on the same inputs; (2) whether the two signals exhibit
similar relationships with correctness; and (3) whether the magnitude of
local--global disagreement is associated with sampling instability at the
question level. We additionally test whether the latter relationship persists
when confidence disagreement and sampling instability are estimated from
non-overlapping stochastic samples.

All analyses use a locally served
\texttt{llama3.3:70b-q4\_H\_K}~\citep{grattafiori2024llama3} language model
accessed through the Ollama API~\citep{ollama2024} with
\texttt{logprobs=true}. Our instantiation required approximately 48 GB of
RAM/VRAM and was run on an M2 Ultra Mac Studio with 64 GB of RAM. No GPU
cluster or proprietary API was used.

\subsection{Datasets}
\label{sec:setup:datasets}

The primary analysis uses 100 questions from
\textbf{MMLU}~\citep{hendrycks2021mmlu}
(\texttt{high\_school\_us\_history}) and 100 questions from
\textbf{ARC Challenge}~\citep{clark2018arc}. Both benchmarks use constrained
multiple-choice answer spaces, allowing token-level probabilities and sampled
answer frequencies to be compared at a common answer level.

This restriction is deliberate. In open-ended generation, semantically
equivalent answers can differ in surface form, making answer-frequency
estimates dependent on an additional semantic-equivalence procedure. The
constrained answer spaces of MMLU and ARC isolate the relationship between
local token confidence and sampled answer concentration without introducing
this additional source of variation.

One ARC question lacked a valid local-confidence measurement and is excluded
from analyses requiring both confidence signals, yielding $n=100$ for MMLU
and $n=99$ for ARC.

Prompts follow the same zero-shot multiple-choice format across the primary
analyses.

\subsection{Confidence Signals}
\label{sec:setup:signals}

For each question, we estimate local and global confidence using separate
readouts of the same autoregressive predictive process.

\paragraph{Local confidence.}
Local confidence is the probability assigned to the greedy-selected answer
token,
\begin{equation}
\clocal(x)
=
p_\theta(a^* \mid x),
\qquad
a^*
=
\operatorname*{arg\,max}_{a}
p_\theta(a\mid x).
\end{equation}
The probability is obtained from the model's token-level log-probability
output using a separate greedy call at $T=0$.

\paragraph{Global confidence.}
Global confidence is the empirical frequency of the modal answer under
repeated stochastic sampling,
\begin{equation}
\cglobal(x)
=
\frac{1}{N}
\sum_{i=1}^{N}
\mathbf{1}[a_i=\hat{a}],
\qquad
\hat{a}
=
\operatorname*{arg\,max}_{a}
\sum_{i=1}^{N}
\mathbf{1}[a_i=a],
\end{equation}
where $a_i$ denotes the answer obtained from the $i$th sample.

For the primary analysis, $\cglobal$ is estimated from 60 samples at
$T=0.7$. These comprise two existing 30-draw batches generated using the
same prompt and sampling-temperature configuration. Pooling the batches
provides the primary estimate of the sampled answer distribution.

\subsection{Sampling Instability}
\label{sec:setup:instability}

Repeated stochastic generation induces an empirical answer distribution
$\hat{P}(a\mid x)$. We characterize question-level sampling instability using
four complementary summaries of this distribution: Shannon entropy,
\begin{equation}
H(x)
=
-\sum_a
\hat{P}(a\mid x)
\log \hat{P}(a\mid x),
\end{equation}
the number of distinct sampled answers, the empirical probability of the
modal answer, and a binary instability indicator equal to one when more than
one distinct answer is observed.

The primary instability measurements are estimated from 30 samples per
question at $T=0.7$.

\subsection{Primary Analyses}
\label{sec:setup:metrics}

\paragraph{Local--global agreement.}
We measure agreement between the two confidence signals using Pearson
correlation $r(\clocal,\cglobal)$. Question-level disagreement is defined as
the absolute confidence gap
\begin{equation}
D(x)
=
|\clocal(x)-\cglobal(x)|.
\end{equation}
We additionally report the mean gap and the fraction of questions for which
$D(x)$ exceeds $0.1$ and $0.2$.

\paragraph{Confidence--correctness association.}
For each confidence signal, we measure its Pearson correlation with binary
answer correctness. These quantities compare how strongly the two confidence
readouts are associated with correctness; they are not interpreted as
complete measures of probabilistic calibration.

\paragraph{Confidence disagreement and sampling instability.}
The central question-level analysis tests whether $D(x)$ is associated with
sampling instability. Because the instability measures are non-Gaussian and
contain many tied observations, we use Spearman rank correlation for
continuous and ordinal measures. For the binary instability indicator, stable
and unstable questions are compared using the Mann--Whitney $U$ test and
rank-biserial effect size. Multiple tests within this analysis are corrected
using the Benjamini--Hochberg procedure.

As a companion analysis, we test the association between $\clocal(x)$ alone
and the same instability measures. Because $\clocal$ is obtained from a
separate greedy log-probability call, it does not share stochastic samples
with the instability estimates. This analysis tests whether the
gap--instability relationship can be reduced to a simpler relationship
between low local confidence and unstable sampling.

\subsection{Disjoint-Sample Robustness}
\label{sec:setup:robustness}

The primary 60-sample estimate of $\cglobal$ includes the 30 baseline draws
used to estimate sampling instability. Consequently, the primary estimates
of $D(x)$ and sampling instability partially share finite-sample variation.

We directly test whether this overlap accounts for the observed association
using disjoint sampling subsets. For each question, the 30 baseline
stochastic draws are randomly divided into two non-overlapping sets of 15.
One subset is used to estimate $\cglobal$ and hence $D(x)$, while the
complementary subset is used to estimate sampling entropy. We repeat this
procedure across 200 random partitions and report the distribution of the
resulting question-level Spearman correlations.

This analysis removes shared Monte Carlo draws between the two estimates. It
does not make global confidence and sampling entropy conceptually independent:
both remain summaries of the same underlying answer distribution. It
therefore tests robustness to shared finite-sample variation rather than
causal independence.

\subsection{Experimental Overview}
\label{sec:setup:experiments}

Table~\ref{tab:experiments} summarizes the empirical design. Rather than
treating each derived statistic as a separate experiment, the analyses are
organized around the measurements required to test the three primary
questions.

\begin{table}[t]
  \centering
  \caption{Summary of the primary empirical analyses.}
  \label{tab:experiments}
  \small
  \begin{adjustbox}{max width=\linewidth}
  \begin{tabular}{llll}
    \toprule
    \textbf{Analysis}
      & \textbf{Datasets}
      & \textbf{Sampling}
      & \textbf{Purpose} \\
    \midrule

    Local--global confidence
      & MMLU, ARC
      & Greedy + $60\times T{=}0.7$
      & Confidence agreement and correctness \\

    Sampling instability
      & MMLU, ARC
      & $30\times T{=}0.7$
      & Question-level output variability \\

    Gap--instability
      & MMLU, ARC
      & Combined
      & Question-level association \\

    Disjoint robustness
      & MMLU, ARC
      & $15/15$, 200 splits
      & Remove shared sampling variation \\

    \bottomrule
  \end{tabular}
  \end{adjustbox}
\end{table}

The first two analyses construct the confidence and instability measurements.
The third provides the central test of whether disagreement between confidence
readouts tracks sampling instability. The fourth tests whether this
relationship persists when the two quantities are estimated from
non-overlapping stochastic draws.

\section{Results}
\label{sec:results}

\subsection{Local and Global Confidence Show Weak Agreement}
\label{sec:results:agreement}

\begin{table}[t]
  \centering
  \caption{Local--global confidence agreement and association with correctness.
    MMLU: 100 questions; ARC: 99 questions (one question lacked a valid
    local-confidence measurement). Global confidence is estimated from
    60 pooled samples at $T=0.7$. Pearson $r$ values are reported with
    95\% CIs obtained via Fisher $z$-transformation. Gap is
    $D(x)=|\clocal(x)-\cglobal(x)|$.}
  \label{tab:confidence}
  \small
  \begin{adjustbox}{max width=\linewidth}
  \begin{tabular}{l p{3.2cm} p{3.2cm}}
    \toprule
    \textbf{Metric}
      & \textbf{MMLU} ($n{=}100$)
      & \textbf{ARC} ($n{=}99$) \\
    \midrule
    $r(\clocal,\cglobal)$
      & $0.086$ {\scriptsize $[-0.112,\ 0.278]$}
      & $0.175$ {\scriptsize $[-0.024,\ 0.361]$} \\[2pt]
    $r(\clocal,\mathrm{correct})$
      & $0.102$ {\scriptsize $[-0.096,\ 0.293]$}
      & $0.072$ {\scriptsize $[-0.128,\ 0.266]$} \\[2pt]
    $r(\cglobal,\mathrm{correct})$
      & $0.494$ {\scriptsize $[0.329,\ 0.627]$}
      & $0.374$ {\scriptsize $[0.189,\ 0.530]$} \\[4pt]
    Mean absolute gap
      & $0.080$
      & $0.045$ \\[2pt]
    $f_{\mathrm{gap}>0.10}$
      & $23.0\%$ {\scriptsize $[14.8\%,\ 31.2\%]$}
      & $15.2\%$ {\scriptsize $[8.1\%,\ 22.2\%]$} \\[2pt]
    $f_{\mathrm{gap}>0.20}$
      & $19.0\%$ {\scriptsize $[11.3\%,\ 26.7\%]$}
      & $10.1\%$ {\scriptsize $[4.2\%,\ 16.0\%]$} \\[4pt]
    Mean $\clocal$
      & $0.921$
      & $0.962$ \\[2pt]
    Mean $\cglobal$
      & $0.993$
      & $0.981$ \\
    \bottomrule
  \end{tabular}
  \end{adjustbox}
\end{table}

Local and global confidence show weak empirical agreement
(Table~\ref{tab:confidence}). Their Pearson correlation is $r=0.086$
(95\%\,CI $[-0.112,\,0.278]$) on MMLU and $r=0.175$
(95\%\,CI $[-0.024,\,0.361]$) on ARC.

The disagreement is also visible at the question level. The mean absolute
gap is $0.080$ on MMLU and $0.045$ on ARC. A gap larger than $0.10$
occurs for 23.0\% of MMLU questions and 15.2\% of ARC questions, while
gaps larger than $0.20$ occur for 19.0\% and 10.1\%, respectively.

These results quantify the extent to which the two confidence
operationalizations differ on the same inputs. As established in
Section~\ref{sec:localglobal}, such disagreement is compatible with their
definitions as distinct summaries of the autoregressive predictive process;
the empirical result here concerns its magnitude and frequency.

\subsection{Confidence Signals Differ in Their Association with Correctness}
\label{sec:results:correctness}

The two confidence signals also differ in their relationship with answer
correctness. Global confidence shows a moderate positive association with
correctness on both datasets:
$r=0.494$ (95\%\,CI $[0.329,\,0.627]$) on MMLU and
$r=0.374$ (95\%\,CI $[0.189,\,0.530]$) on ARC.

By comparison, local confidence shows little association with correctness:
$r=0.102$ (95\%\,CI $[-0.096,\,0.293]$) on MMLU and
$r=0.072$ (95\%\,CI $[-0.128,\,0.266]$) on ARC.

Thus, in these experiments, the two confidence measurements are not only
weakly associated with one another but also differ in the information they
provide about whether the model's selected answer is correct. These
correlations characterize confidence--correctness association and should
not be interpreted as complete measures of probabilistic calibration.

\subsection{Sampling Instability Differs Between MMLU and ARC}
\label{sec:results:stochasticity}

Before examining the relationship between confidence disagreement and
sampling instability, we characterize the amount of instability available
at the question level. Under the baseline $T=0.7$, $N=30$ condition,
MMLU exhibits mean answer entropy of $0.025$, with 4\% of questions
producing more than one distinct answer. ARC exhibits greater variability,
with mean entropy of $0.071$ and approximately 11\% of questions producing
more than one answer.

Sampling instability is therefore sparse on both datasets but particularly
rare on MMLU. This difference is important for interpreting the subsequent
question-level analysis: ARC provides substantially more variation with
which to identify an association between confidence disagreement and
sampling instability, whereas MMLU operates close to a floor in the
instability measures.

\subsection{Confidence Disagreement Tracks Sampling Instability}
\label{sec:results:gap_instability}

We next test whether question-level local--global confidence disagreement
is associated with question-level sampling instability.
Table~\ref{tab:gap_instability} reports the corresponding associations and
the companion analysis using local confidence alone.

\begin{table}[t]
  \centering
  \caption{Question-level relationship between confidence measurements and
  sampling instability. Correlations are Spearman $\rho$. Binary
  stable--unstable comparisons report rank-biserial effect sizes.
  Reported significance values are Benjamini--Hochberg adjusted where
  available. Disjoint-sample estimates use non-overlapping stochastic
  samples for confidence disagreement and entropy.}
  \label{tab:gap_instability}
  \small
  \begin{adjustbox}{max width=\linewidth}
  \begin{tabular}{lcc}
    \toprule
    \textbf{Association}
      & \textbf{MMLU}
      & \textbf{ARC} \\
    \midrule

    $D(x)$ vs.\ entropy
      & $\rho=0.215$
      & $\rho=0.418^{***}$ \\[2pt]

    $D(x)$ vs.\ distinct answers
      & ---
      & $\rho=0.41^{***}$ \\[2pt]

    $D(x)$ vs.\ modal-answer probability
      & ---
      & $\rho=-0.42^{***}$ \\[4pt]

    Stable vs.\ unstable $D(x)$
      & ---
      & $r_{\mathrm{rb}}=0.76^{***}$ \\[4pt]

    $\clocal$ vs.\ instability
      & $\rho=-0.09$ ($p=0.48$)
      & $\rho=-0.31$ ($p_{\mathrm{BH}}=0.003$) \\[2pt]

    Stable vs.\ unstable $\clocal$
      & ---
      & $r_{\mathrm{rb}}=-0.56$ \\

    \midrule
    \multicolumn{3}{l}{\textit{Disjoint-sample robustness:
    $D(x)$ vs.\ entropy}} \\[2pt]

    Pooled estimate
      & $\rho=0.215$
      & $\rho=0.418$ \\[2pt]

    Mean disjoint-split estimate
      & $\rho=0.214$
      & $\rho=0.397$ \\[-1pt]

    2.5--97.5\% split range
      & $[0.184,\ 0.235]$
      & $[0.354,\ 0.422]$ \\

    \bottomrule
  \end{tabular}
  \end{adjustbox}

  \vspace{2pt}
  \begin{minipage}{0.96\linewidth}
  \scriptsize
  $^{***}$BH-adjusted $p<0.001$. The disjoint-sample estimates summarize
  200 random non-overlapping 15/15 partitions of the baseline stochastic
  samples. The reported split ranges are the 2.5th and 97.5th percentiles
  across partitions and are not confidence intervals.
  \end{minipage}
\end{table}

On ARC, larger confidence gaps are consistently associated with greater
sampling instability. The absolute gap $D(x)$ is positively associated
with answer entropy (Spearman $\rho=0.418$) and the number of distinct
sampled answers ($\rho=0.41$), and negatively associated with modal-answer
probability ($\rho=-0.42$). These associations remain significant after
Benjamini--Hochberg correction (adjusted $p<0.001$).

The stable--unstable comparison yields the same pattern. ARC questions
that produce more than one answer under repeated sampling have a median
absolute confidence gap of $0.12$, compared with $0.00$ among stable
questions, with a rank-biserial effect size of $0.76$.

The corresponding gap--entropy relationship on MMLU is weaker
($\rho=0.215$). Only four of 100 MMLU questions are unstable under the
baseline sampling condition, leaving comparatively little question-level
variation with which to estimate relationships involving sampling
instability.

The primary global-confidence estimate uses 60 samples, including the 30
baseline samples from which the primary instability measures are computed.
We therefore repeat the gap--entropy analysis using non-overlapping
stochastic samples. Across 200 random 15/15 partitions, the ARC association
remains close to the pooled estimate: $\rho=0.418$ in the pooled analysis
and mean $\rho=0.397$ across disjoint splits, with a 2.5--97.5\% split
range of $[0.354,\,0.422]$. MMLU is similarly stable, with $\rho=0.215$
in the pooled analysis and mean $\rho=0.214$ across disjoint splits, with
a split range of $[0.184,\,0.235]$.

The persistence of the relationship under disjoint estimation indicates
that shared finite-sample variation between the global-confidence and
instability estimates does not account for the observed association.

As a companion analysis, we examine local confidence independently of the
local--global gap. On ARC, higher local confidence is associated with lower
sampling instability (Spearman $\rho=-0.31$, BH-adjusted $p=0.003$), and
unstable questions exhibit lower local confidence (rank-biserial effect
size $=-0.56$). On MMLU, local confidence shows little relationship with
instability ($\rho=-0.09$, $p=0.48$). Because $\clocal$ is obtained from a
separate greedy log-probability call, this comparison does not share the
stochastic samples used to estimate instability.

Taken together, the ARC results show that questions exhibiting greater
local--global confidence disagreement also exhibit greater variability
under repeated sampling. The relationship remains essentially unchanged
when disagreement and instability are estimated from non-overlapping
stochastic draws. The weaker MMLU relationship occurs in a regime where
sampling instability is rare.

\section{Discussion}
\label{sec:discussion}

\paragraph{Confidence depends on how the predictive distribution is summarized.}
The central empirical result is that local token confidence and global answer concentration behave as distinct confidence measurements, even when obtained from the same model on the same inputs. Their correlation is weak on both MMLU and ARC, and non-trivial local--global gaps occur for a substantial fraction of questions. This disagreement is not unexpected from first principles: the two quantities summarize different aspects of an autoregressive predictive process. The empirical contribution is therefore not the existence of a possible mismatch, but its magnitude and whether that mismatch is systematically related to observable model behavior.

\paragraph{Local and global confidence carry different information about correctness.}
The two confidence signals also differ markedly in their association with answer correctness. Global confidence is moderately associated with correctness on both MMLU and ARC, whereas local confidence shows little relationship with correctness. This asymmetry indicates that the probability assigned to a selected answer token and the concentration of probability mass across sampled answers should not be treated as interchangeable measures of model confidence.

One likely contributor to this difference is the operational definition of local confidence used here. Because $\clocal$ is the probability of the greedy-selected answer token, it is conditioned on a locally preferred continuation and is frequently concentrated near high values. By contrast, $\cglobal$ summarizes concentration across complete sampled answers. The two measurements consequently retain different information about the predictive process. We do not interpret the confidence--correctness correlations reported here as complete measures of probabilistic calibration.

\paragraph{Confidence disagreement tracks sampling instability.}
The question-level analysis provides the strongest new evidence in the study. On ARC, larger local--global confidence gaps are associated with higher answer entropy, more distinct sampled answers, lower modal-answer concentration, and a greater likelihood of observing more than one answer under repeated sampling. The stable--unstable comparison yields the same pattern: unstable questions exhibit substantially larger confidence gaps.

This relationship is not reducible to a generic effect of low local confidence. On ARC, local confidence alone is negatively associated with instability ($\rho=-0.31$), whereas the local--global gap exhibits a stronger positive association with answer entropy ($\rho=0.42$). The gap therefore captures the relationship between a locally concentrated prediction and the concentration of the resulting answer distribution, rather than simply identifying questions for which the local prediction is uncertain.

The corresponding relationship is substantially weaker on MMLU. However, MMLU operates in a near-floor instability regime, with only four of 100 questions producing more than one answer under the baseline condition. The contrast between ARC and MMLU therefore indicates that the observability of the gap--instability relationship depends on there being sufficient variation in sampling behavior.

\paragraph{The relationship is robust to disjoint sampling but remains diagnostic.}
The primary analysis estimates global confidence from 60 pooled samples, 30 of which are also used to estimate the baseline instability measures. This creates shared finite-sample variation between the two estimates. The disjoint-sample analysis directly tests whether this overlap accounts for the observed relationship.

It does not. On ARC, the gap--entropy correlation is $\rho=0.418$ in the pooled analysis and averages $\rho=0.397$ across 200 disjoint-sample splits (2.5-97.5\% split interval $[0.354,\,0.422]$). On MMLU, the corresponding estimates are $\rho=0.215$ and $\rho=0.214$ ($[0.184,\,0.235]$). The persistence of the association under non-overlapping sampling therefore indicates that it is not explained by shared Monte Carlo variation.

This robustness result does not, however, establish causal direction. Global confidence and sampling entropy remain different summaries of the same underlying answer distribution even when estimated from independent draws. Their association can therefore arise because both reflect latent variation in that distribution. The present experiments establish that local--global disagreement tracks sampling instability across questions; they do not establish whether disagreement contributes to instability, instability enlarges the observed confidence gap, or both reflect a common property of the predictive distribution.

\paragraph{Implications for evaluation, oversight, and alignment.}
These results have implications beyond the comparison of uncertainty metrics.
Confidence increasingly serves as an interface between model predictions and
downstream mechanisms for evaluation, abstention, human review, feedback, and
control. Such mechanisms implicitly require a choice about which property of
the predictive process is being treated as confidence.

The present results show that this choice can be consequential. Local token
confidence and sampled answer concentration can assign substantially different
confidence to the same input, exhibit different relationships with correctness,
and disagree most strongly on questions that also exhibit greater sampling
instability. A downstream mechanism conditioned on one confidence readout may
therefore receive materially different information from one conditioned on
another, even when both are described as responding to ``model confidence.''

This constitutes a measurement problem for confidence-based alignment and
oversight rather than evidence about alignment performance itself. Before a
confidence signal can be calibrated against correctness, used to trigger
deferral, or incorporated into human or automated feedback, the quantity being
measured must be specified. Our results suggest that confidence should not be
assumed to provide a representation-independent scalar summary of the model's
predictive state.

\paragraph{Future work.}
The immediate next step is to determine whether the observed relationships generalize across model families, scales, and training regimes. Larger benchmark samples are also needed to characterize low-instability regimes such as MMLU more precisely.

A second direction is to test robustness to calibration of the local confidence signal. In particular, post-hoc temperature scaling would test whether the observed local--global asymmetry and its relationship with sampling instability persist after rescaling local token probabilities. Because such calibration was not performed here, the present results do not establish invariance to this transformation.

Finally, extending the analysis beyond constrained-answer tasks will require explicit treatment of semantic equivalence, since surface-form diversity can otherwise be confounded with genuine answer-level instability.

\section{Conclusion}

Local token confidence and global answer concentration provide empirically
distinct views of the same autoregressive predictive process. In our
experiments, the two signals show weak agreement, differ substantially in
their association with correctness, and, on ARC, their disagreement is
systematically associated with sampling instability at the question level.
This association persists when confidence disagreement and instability are
estimated from disjoint stochastic samples, indicating that it is not
explained by shared finite-sample variation.

These results support treating confidence as an explicitly defined
measurement rather than as a single intrinsic scalar property of a language
model. In particular, token-level confidence and sampling-derived answer
concentration should not be assumed to be interchangeable. The local--global
confidence gap provides a simple diagnostic for identifying inputs on which
these two readouts diverge and sampling behavior is less stable.

The present results establish association rather than causal direction, and
determining why confidence disagreement and sampling instability co-occur
remains an open empirical question. More broadly, however, the findings expose
a measurement problem for systems that use confidence as an input to
evaluation, oversight, or control. If different readouts of the same predictive
process disagree and carry different information about correctness and
stability, then ``model confidence'' cannot be treated as an operationally
unique quantity without specifying how it is measured. Establishing which
confidence representation is appropriate for a given downstream objective is
therefore a prerequisite for using confidence reliably in such systems.

\paragraph{Limitations}

The study has several limitations. First, all experiments use a single
quantized Llama~3.3 70B model, so the magnitude of the observed effects may
differ across model families, scales, and training regimes. Second, the
primary local--global analysis uses approximately 100 questions from MMLU
and ARC. This limits precision, particularly in low-instability regimes such
as MMLU, where only four questions exhibit sampling instability under the
baseline condition.

Third, the local-confidence measure used here is specifically the probability
assigned to the greedy-selected answer token. The results therefore
characterize this operationalization rather than token-level uncertainty
measures in general. Global confidence is estimated through repeated sampling
and consequently depends on the sampling configuration and finite sample
budget. We also do not test whether the observed relationships persist after
post-hoc calibration, such as temperature scaling of the local confidence
signal.

Fourth, the primary comparison is restricted to multiple-choice tasks. This
is deliberate: constrained answer spaces reduce ambiguity from synonyms,
paraphrases, and surface-form variation, allowing local token probabilities
and sampled answer frequencies to be compared at a common answer level.
Extending the analysis to open-ended generation will require explicit
treatment of semantic equivalence.

Finally, the gap--instability analysis is observational. In the primary
analysis, global confidence is estimated from 60 pooled samples, 30 of which
are also used to estimate baseline sampling instability. We therefore repeat
the analysis across 200 disjoint splits, estimating the two quantities from
non-overlapping stochastic draws; the gap--entropy association remains
essentially unchanged. This rules out shared finite-sample variation as an
explanation for the observed relationship. However, global confidence and
sampling instability remain summaries of the same underlying answer
distribution. The analysis therefore supports a diagnostic association, not
a causal interpretation of the relationship.

\paragraph{Broader Impacts}

The study concerns measurement and evaluation rather than the development of
new model capabilities. Its primary implication is that confidence estimates
should be reported together with a clear definition of how they are obtained.
Treating distinct confidence signals as interchangeable may obscure meaningful
differences in model behavior, particularly when confidence is used to support
downstream decisions.

Comparing multiple confidence readouts may therefore improve diagnostic
evaluation of model reliability. At the same time, disagreement between
confidence measures should not itself be interpreted as a calibrated estimate
of risk or as evidence of a particular causal mechanism. In higher-stakes
applications, any such diagnostic should be validated for the specific model,
task, and deployment setting before being used operationally.

\bibliographystyle{plainnat}
\bibliography{references}

@inproceedings{guo2017calibration,
  title     = {On Calibration of Modern Neural Networks},
  author    = {Guo, Chuan and Pleiss, Geoff and Sun, Yu and Weinberger, Kilian Q.},
  booktitle = {Proceedings of the 34th International Conference on Machine Learning (ICML)},
  year      = {2017},
  url       = {https://arxiv.org/abs/1706.04599}
}

@inproceedings{niculescu2005predicting,
  title     = {Predicting Good Probabilities with Supervised Learning},
  author    = {Niculescu-Mizil, Alexandru and Caruana, Rich},
  booktitle = {Proceedings of the 22nd International Conference on Machine Learning (ICML)},
  year      = {2005},
  doi       = {10.1145/1102351.1102430},
  url       = {https://dl.acm.org/doi/10.1145/1102351.1102430}
}

@article{jiang2021know,
  title   = {How Can We Know When Language Models Know? {On} the Calibration of Language Models
             for Question Answering},
  author  = {Jiang, Zhengbao and Araki, Jun and Ding, Haibo and Neubig, Graham},
  journal = {Transactions of the Association for Computational Linguistics},
  volume  = {9},
  year    = {2021},
  url     = {https://arxiv.org/abs/2012.00955}
}

@inproceedings{zhao2023slic,
  title     = {Calibrating Sequence likelihood Improves Conditional Language Generation},
  author    = {Zhao, Yao and Khalman, Misha and Joshi, Rishabh and Narayan, Shashi
               and Saleh, Mohammad and Liu, Peter J.},
  booktitle = {Proceedings of the 11th International Conference on Learning Representations
               (ICLR)},
  year      = {2023},
  url       = {https://arxiv.org/abs/2210.00045}
}

@inproceedings{kuhn2023semantic,
  title     = {Semantic Uncertainty: Linguistic Invariances for Uncertainty Estimation
               in Natural Language Generation},
  author    = {Kuhn, Lorenz and Gal, Yarin and Farquhar, Sebastian},
  booktitle = {Proceedings of the 11th International Conference on Learning Representations
               (ICLR)},
  year      = {2023},
  url       = {https://arxiv.org/abs/2302.09664}
}

@inproceedings{xiong2024llms,
  title     = {Can {LLMs} Express Their Uncertainty? {An} Empirical Evaluation of Confidence
               Elicitation in {LLMs}},
  author    = {Xiong, Miao and Hu, Zhiyuan and Lu, Xinyang and Li, Yifei and Fu, Jie
               and He, Junxian and Hooi, Bryan},
  booktitle = {Proceedings of the 12th International Conference on Learning Representations
               (ICLR)},
  year      = {2024},
  url       = {https://arxiv.org/abs/2306.13063}
}

@misc{kadavath2022language,
  title  = {Language Models (Mostly) Know What They Know},
  author = {Kadavath, Saurav and Conerly, Tom and Askell, Amanda and Henighan, Tom
            and others},
  year   = {2022},
  note   = {arXiv preprint arXiv:2207.05221. Anthropic technical report.},
  url    = {https://arxiv.org/abs/2207.05221}
}

@article{lin2022teaching,
  title   = {Teaching Models to Express Their Uncertainty in Words},
  author  = {Lin, Stephanie and Hilton, Jacob and Evans, Owain},
  journal = {Transactions on Machine Learning Research (TMLR)},
  year    = {2022},
  url     = {https://arxiv.org/abs/2205.14334}
}

@inproceedings{tian2023calibration,
  title     = {Just Ask for Calibration: Strategies for Eliciting Calibrated Confidence Scores
               from Language Models Fine-Tuned with Human Feedback},
  author    = {Tian, Katherine and Mitchell, Eric and Zhou, Allan and Sharma, Archit
               and Rafailov, Rafael and Yao, Huaxiu and Finn, Chelsea and Manning, Christopher D.},
  booktitle = {Proceedings of the 2023 Conference on Empirical Methods in Natural Language
               Processing (EMNLP)},
  year      = {2023},
  url       = {https://arxiv.org/abs/2305.14975}
}

@inproceedings{kumar2024confidence,
  title     = {Confidence Under the Hood: An Investigation into the Confidence-Probability
               Alignment in Large Language Models},
  author    = {Kumar, Abhishek and Morabito, Robert and Umbet, Sanzhar and Kabbara, Jad
               and Emami, Ali},
  booktitle = {Proceedings of the 62nd Annual Meeting of the Association for Computational
               Linguistics (ACL)},
  year      = {2024},
  url       = {https://arxiv.org/abs/2405.16282}
}

@inproceedings{wang2023selfconsistency,
  title     = {Self-Consistency Improves Chain of Thought Reasoning in Language Models},
  author    = {Wang, Xuezhi and Wei, Jason and Schuurmans, Dale and Le, Quoc and Chi, Ed
               and Narang, Sharan and Chowdhery, Aakanksha and Zhou, Denny},
  booktitle = {Proceedings of the 11th International Conference on Learning Representations
               (ICLR)},
  year      = {2023},
  url       = {https://arxiv.org/abs/2203.11171}
}

@misc{chen2023universal,
  title  = {Universal Self-Consistency for Large Language Model Generation},
  author = {Chen, Xinyun and Aksitov, Renat and Alon, Uri and Ren, Jie and Xiao, Kefan
            and Yin, Pengcheng and Prakash, Sushant and Sutton, Charles and Wang, Xuezhi
            and Zhou, Denny},
  year   = {2023},
  note   = {arXiv preprint arXiv:2311.17311.},
  url    = {https://arxiv.org/abs/2311.17311}
}

@inproceedings{burns2023discovering,
  title     = {Discovering Latent Knowledge in Language Models Without Supervision},
  author    = {Burns, Collin and Ye, Haotian and Klein, Dan and Steinhardt, Jacob},
  booktitle = {Proceedings of the 11th International Conference on Learning Representations
               (ICLR)},
  year      = {2023},
  url       = {https://arxiv.org/abs/2212.03827}
}

@misc{marks2023geometry,
  title  = {The Geometry of Truth: Emergent Linear Structure in Large Language Model
            Representations of True/{False} Datasets},
  author = {Marks, Samuel and Tegmark, Max},
  year   = {2023},
  note   = {arXiv preprint arXiv:2310.06824},
  url    = {https://arxiv.org/abs/2310.06824}
}

@inproceedings{zhao2021calibrate,
  title     = {Calibrate Before Use: Improving Few-Shot Performance of Language Models},
  author    = {Zhao, Tony Z. and Wallace, Eric and Feng, Shi and Klein, Dan and Singh, Sameer},
  booktitle = {Proceedings of the 38th International Conference on Machine Learning (ICML)},
  year      = {2021},
  url       = {https://arxiv.org/abs/2102.09690}
}

@inproceedings{sclar2024quantifying,
  title     = {Quantifying Language Models' Sensitivity to Spurious Features in Prompt Design
               or: How {I} learned to start worrying about prompt formatting},
  author    = {Sclar, Melanie and Choi, Yejin and Tsvetkov, Yulia and Suhr, Alane},
  booktitle = {Proceedings of the 12th International Conference on Learning Representations
               (ICLR)},
  year      = {2024},
  url       = {https://arxiv.org/abs/2310.11324}
}

@inproceedings{malinin2021uncertainty,
  title     = {Uncertainty Estimation in Autoregressive Structured Prediction},
  author    = {Malinin, Andrey and Gales, Mark},
  booktitle = {Proceedings of the 9th International Conference on Learning Representations
               (ICLR)},
  year      = {2021},
  url       = {https://arxiv.org/abs/2002.07650}
}

@inproceedings{ren2023robots,
  title     = {Robots That Ask For Help: Uncertainty Alignment for Large Language Model Planners},
  author    = {Ren, Allen Z. and Dixit, Anushri and Bodrova, Alexandra and Singh, Sumeet
               and Tu, Stephen and Brown, Noah and Xu, Peng and Takayama, Leila and Xia, Fei
               and Varley, Jake and Xu, Zhenjia and Sadigh, Dorsa and Zeng, Andy
               and Majumdar, Anirudha},
  booktitle = {Proceedings of the 7th Conference on Robot Learning (CoRL)},
  year      = {2023},
  url       = {https://arxiv.org/abs/2307.01928}
}

@inproceedings{hendrycks2021mmlu,
  title     = {Measuring Massive Multitask Language Understanding},
  author    = {Hendrycks, Dan and Burns, Collin and Basart, Steven and Zou, Andy
               and Mazeika, Mantas and Song, Dawn and Steinhardt, Jacob},
  booktitle = {Proceedings of the 9th International Conference on Learning Representations
               (ICLR)},
  year      = {2021},
  url       = {https://arxiv.org/abs/2009.03300}
}

@misc{clark2018arc,
  title  = {Think you have Solved Question Answering? {Try} {ARC}, the {AI2} Reasoning Challenge},
  author = {Clark, Peter and Cowhey, Isaac and Etzioni, Oren and Khot, Tushar
            and Sabharwal, Ashish and Schoenick, Carissa and Tafjord, Oyvind},
  year   = {2018},
  note   = {arXiv preprint arXiv:1803.05457.},
  url    = {https://arxiv.org/abs/1803.05457}
}

@misc{grattafiori2024llama3,
  title  = {The {Llama} 3 Herd of Models},
  author = {Grattafiori, Aaron and Dubey, Abhimanyu and Jauhri, Abhinav and Pandey, Abhinav
            and Kadian, Abhishek and Al-Dahle, Ahmad and others},
  year   = {2024},
  note   = {arXiv preprint arXiv:2407.21783. The specific model used in experiments is
            Llama~3.3 70B Instruct, quantized to 4-bit precision
            (\texttt{Q4\_K\_M} GGUF format) and served locally via Ollama as
            \texttt{llama3.3:70b-instruct-q4\_K\_M}. Model card:
            \url{https://huggingface.co/meta-llama/Llama-3.3-70B-Instruct}.},
  url    = {https://arxiv.org/abs/2407.21783}
}

@misc{ollama2024,
  title  = {Ollama},
  author = {{Ollama contributors}},
  year   = {2024},
  note   = {Open-source runtime for serving large language models locally.
            The \texttt{llama3.3:70b-instruct-q4\_K\_M} model was served via Ollama's
            OpenAI-compatible API at \texttt{http://localhost:11434}.},
  url    = {https://github.com/ollama/ollama}
}


\end{document}